\documentclass[preprint,12pt]{elsarticle}

\usepackage{amsmath,amsfonts}
\usepackage{algorithmic}
\usepackage{algorithm}
\usepackage{array}
\usepackage[caption=false,font=scriptsize,labelfont=sf,textfont=sf]{subfig}
\usepackage{textcomp}
\usepackage{stfloats}
\usepackage{url}
\usepackage{verbatim}
\usepackage{graphicx}
\usepackage{cite}

\usepackage{newtxmath}
\usepackage{multirow}
\usepackage{booktabs}
\usepackage{CJK}
\usepackage{xpinyin}
\usepackage{color, xcolor}

\newcommand{\methodName}{ProTEC}

\journal{Expert Systems with Applications}

\begin{document}

\begin{frontmatter}



\title{Correct Like Humans: Progressive Learning Framework for Chinese Text Error Correction}

\author[thu]{Yinghui Li\fnref{cor1}}
\author[thu]{Shirong Ma\fnref{cor1}} 
\author[scau]{Shaoshen Chen\fnref{cor1}} 
\author[thu]{Haojing Huang} 
\author[thu]{Shulin Huang} 
\author[thu]{Yangning Li}
\author[thu]{Hai-Tao Zheng\corref{cor2}} 
\author[sysu]{Ying Shen\corref{cor2}}
\fntext[cor1]{indicates equal contribution. }
\cortext[cor2]{Corresponding authors.}
\fntext[label3]{E-mails: liyinghu20@mails.tsinghua.edu.cn, masr21@mails.tsinghua.edu.cn, chen77@stu.scau.edu.cn, hhj23@mails.tsinghua.edu.cn, \\ sl-huang21@mails.tsinghua.edu.cn, yn-li23@mails.tsinghua.edu.cn, zheng.haitao@sz.tsinghua.edu.cn, sheny76@mail.sysu.edu.cn}

\affiliation[thu]{organization={Tsinghua Shenzhen International Graduate School, Tsinghua University},
            city={Shenzhen},
            postcode={518055}, 
            state={Guangdong},
            country={China}}

\affiliation[scau]{organization={South China Agricultural University},
            city={Guangzhou},
            postcode={510642}, 
            state={Guangdong},
            country={China}}

\affiliation[sysu]{organization={Sun-Yat Sen University},
            city={Guangzhou},
            postcode={510275}, 
            state={Guangdong},
            country={China}}

\begin{abstract}
Chinese Text Error Correction (CTEC) aims to detect and correct errors in the input text, which benefits human daily life and various downstream tasks.
Recent approaches mainly employ Pre-trained Language Models (PLMs) to resolve CTEC.
Although PLMs have achieved remarkable success in CTEC, we argue that previous studies still overlook the importance of human thinking patterns.
To enhance the development of PLMs for CTEC, inspired by humans’ daily error-correcting behavior, we propose a novel model-agnostic progressive learning framework, named \methodName{}, which guides PLMs-based CTEC models to learn to correct like humans.
During the training process, \methodName{} guides the model to learn text error correction by incorporating these sub-tasks into a progressive paradigm. 
During the inference process, the model completes these sub-tasks in turn to generate the correction results.
Extensive experiments and detailed analyses demonstrate the effectiveness and efficiency of our proposed model-agnostic \methodName{} framework.
\end{abstract}


\begin{highlights}
\item Progressive learning with increasing difficulty guides models to correct like humans.
\item Model-agnostic framework can be adapted to various CTEC (including CGEC/CSC) models.
\item Extensive experiments demonstrate the effectiveness and efficiency of ProTEC.
\end{highlights}


\end{frontmatter}


\section{Introduction}
Chinese Text Error Correction (CTEC) aims to detect and correct various types of errors in an input sentence and provide the corresponding correct sentence~\citep{wu2013integrating, wang2020comprehensive}.
As a crucial task, CTEC has attracted increasing attention from NLP researchers due to its broader applications in various NLP and IR downstream tasks~\citep{duan2011online, afli2016using, kubis2019open}.
According to different error types, CTEC is often divided into two types of tasks, namely text-aligned Chinese Spelling Correction (CSC) and text-non-aligned Chinese Grammatical Error Correction (CGEC), and researchers usually study these two tasks separately. 
To promote a more unified development of the CTEC field, in our work, we focus on studying how to handle both types of these two tasks simultaneously. Table~\ref{tab:intro} presents several examples of the CTEC task.

In recent years, methods based on Pre-trained Language Models (PLMs) have become the mainstream of CTEC task~\citep{li2022past, li-etal-2022-improving-chinese, zhang2022mucgec}. Researchers carefully devises pre-training or fine-tuning processes to incorporate task-specific knowledge into PLMs. Existing CTEC approaches can be mainly divided into two categories: Sequence-to-Tagging methods (Seq2Tag), which predict the edit labels corresponding to each position in the input sentence~\citep{omelianchuk2020gector, liang2020bert}, and Sequence-to-Sequence (Seq2Seq) methods, which directly generate the corrected sentence based on the input sentence~\citep{zhao2020maskgec, tang2021chinese}.

\begin{CJK*}{UTF8}{gbsn}

\begin{table}[ht]
\centering
\small
\caption{Examples of Chinese Text Error Correction (CTEC) task.}
\label{tab:intro}
\begin{tabular}{ll} 
\toprule
Input: & \begin{tabular}[c]{@{}l@{}}我听说这个礼拜六你要开一个\textcolor{red}{误}会。\\I heard you are having a \textcolor{red}{misunderstanding} this Saturday.\end{tabular} \\
Output: & \begin{tabular}[c]{@{}l@{}}我听说这个礼拜六你要开一个\textcolor{blue}{舞}会。\\I heard you are having a \textcolor{blue}{dance} this Saturday.\end{tabular} \\
Edit(s): & ``误"~$\rightarrow$~Replace\_``舞" (Phonological Error) \\ 
\midrule
Input: & \begin{tabular}[c]{@{}l@{}}大家害怕你的工厂把自然\textcolor{red}{被}坏。\\People are afraid that your factory will \textcolor{red}{be destroyed} nature.\end{tabular} \\
Output: & \begin{tabular}[c]{@{}l@{}}大家害怕你的工厂把自然\textcolor{blue}{破}坏。\\People are afraid that your factory will \textcolor{blue}{destroy} nature.\end{tabular} \\
Edit(s): & ``被"~$\rightarrow$~Replace\_``破" (Morphological Error) \\ 
\midrule
Input: & \begin{tabular}[c]{@{}l@{}}沙尘暴也是一类空气污染\textcolor{red}{之一}。\\Sandstorm is also \textcolor{red}{one of} a kind of air pollution.\end{tabular} \\
Output: & \begin{tabular}[c]{@{}l@{}}沙尘暴也是一类空气污染。\\Sandstorm is also a kind of air pollution.\end{tabular} \\
Edit(s): & ``之"~$\rightarrow$~Delete; ``一"~$\rightarrow$~Delete \\ 
\midrule
Input: & \begin{tabular}[c]{@{}l@{}}可它的表情是从来没看过的。\\But it has an expression never seen before.\end{tabular} \\
Output: & \begin{tabular}[c]{@{}l@{}}可它的表情是\textcolor{blue}{我}从来没看过的。\\But it has an expression \textcolor{blue}{I have} never seen before.\end{tabular} \\
Edit(s): & ``是" $\rightarrow$~Append\_``我" \\
\bottomrule
\end{tabular}
\end{table}

\end{CJK*}

Although PLMs have achieved remarkable success in CTEC, we argue that previous studies still overlook the importance of human thinking patterns, which hinders further improvement of CTEC performance. As a task closely related to humans' daily lives, the thinking pattern of humans in the text correction process is undoubtedly important. When a person tries to correct an erroneous sentence, he or she would first observe the sentence to \textbf{detect} which positions contain errors. Then, he or she would \textbf{identify} within the context of the sentence to determine the type of error. Finally, he or she would decide what the corresponding erroneous characters, words, or phrases should be \textbf{corrected}.

Inspired by the human thinking pattern in text correction, we propose a novel progressive learning framework for CTEC, named \methodName{}. The framework decomposes the CTEC task into three sub-tasks of varying difficulty levels, allowing CTEC models to progressively learn and do corrections like humans. \methodName{} is a model-agnostic framework and can be applied to various existing CTEC models in both text-aligned and text-non-aligned scenarios (i.e., for errors of CSC and CGEC). Intuitively, we believe that our proposed \methodName{} framework can bring improvements to previous state-of-the-art CTEC models, thanks to the fact that it can guide CTEC models to learn and correct errors from easy to difficult like humans, which is also consistent with the core idea of widely studied curriculum learning~\citep{bengio2009curriculum}.

Specifically, our framework divides the CTEC task into the following three sub-tasks from easy to difficult.
(1) \textbf{Error Detection (Low-Level)}: the CTEC model should detect incorrect positions in a sentence, i.e., judge whether each token in a sentence is erroneous or not. (2) \textbf{Error Type Identification (Mid-Level)}: The model should analyze the error type of each erroneous token. (3) \textbf{Correction Result Generation (High-Level)}: The model generates the final error correction result with restriction of the error type analyzed in the previous sub-task.
During the training process, we employ these three sub-tasks to construct a multi-task training objective, which directs the model to learn the process of text error correction progressively from easy to difficult. During the inference process, the model should complete these three sub-tasks in sequence, gradually narrowing down the range of candidates until the final correction results are generated. Notably, the model leverages the output of previous sub-tasks to perform subsequent sub-tasks. For instance, during the correction result generation step, tokens that have been identified as error-free in the preceding sub-tasks are exempted from correction, thereby mitigating the long-standing problem of over-correction faced by PLMs-based CTEC models.

We evaluate our framework on widely used datasets and extensive experimental results show that \methodName{} can significantly improve the CTEC performance of various models on different datasets, while applying \methodName{} slightly adds only about 5\%-10\% of training and inference time compared to baselines.
This demonstrates the effectiveness and efficiency of our framework.
Our contributions in this paper are summarized as follows:
\begin{itemize}
    \item We present a novel approach by decomposing the CTEC task into multiple sub-tasks of varying difficulty levels and incorporating token-level information on error presence and error types during training and inference to mitigate error propagation and over-correction.
    \item We propose a model-agnostic text correction framework, \methodName{}, which can be adapted to various CTEC models in both text-aligned and text-non-aligned scenarios, facilitating progressive learning for those models and guiding them to correct errors like humans.
    \item We conduct extensive experiments on publicly available CTEC datasets. Experimental results demonstrate the effectiveness and efficiency of \methodName{} in enhancing the performance of various models significantly, with only adding a small amount of training and inference time.
\end{itemize}

\section{Related Work}\label{sec:rw}
In the NLP research community, Chinese Text Error Correction (CTEC) typically involves two tasks: Chinese Spelling Correction (CSC) and Chinese Grammatical Error Correction (CGEC). The CSC task focuses on detecting and correcting spelling errors in text and is a text-aligned task. Spelling errors represents that some characters in a Chinese sentence are written incorrectly as other characters that have similar pronunciation or shape. The CGEC task aims to identify and correct various grammatical errors that violate Chinese language rules, including but not limited to semantic omissions, repetitions, structural confusion, and improper word order and is a text-non-aligned task.
\subsection{Chinese Spelling Correction}
Early works on CSC focus on designing various rules to detect and correct different errors~\citep{chang2015introduction, chu2015ntou}.
Then, statistical machine learning algorithms are applied to resolve CSC task, such as Hidden Markov Model (HMM) and Conditional Random Field (CRF)~\citep{zhang2015hanspeller++, wang2015word}.
In recent years, deep learning-based methods have been the mainstream of CSC. \citet{wang2018hybrid} regards CSC as a sequence labeling task, which utilizes a bidirectional LSTM~\citep{hochreiter1997long} to predict correct characters corresponding to input characters. \citet{wang2019confusionset} proposes a sequence-to-sequence model named confusion set-guided pointer network to generate the corrected sentence for CSC task.
After pre-trained language models (PLMs), e.g. BERT~\citep{devlin2019bert}, were proposed, more and more CSC approaches employ PLMs as their backbones. Soft-Masked BERT~\citep{zhang2020spelling} proposes a cascading model containing two neural networks, a GRU~\citep{chung2014empirical} for error detection and a BERT for error correction. SpellGCN~\citep{cheng2020spellgcn} employs GCN~\citep{kipf2016semi} to model the character similarity in the confusion set. Two-ways~\citep{li2021exploration} identifies weak spots of the model to generate more valuable samples for model training.
Some works pay attention to multi-modal knowledge of characters, and employ it to enhance CSC models. MLM-phonetics~\citep{zhang2021correcting} and DCN~\citep{wang2021dynamic} incorporate phonological information as external knowledge into PLMs to improve model performance. PLOME~\citep{liu2021plome} and REALISE~\citep{xu2021read} leverages phonetics, glyph and semantic information of input characters, and then fuses them to predict final correct characters.
LEAD~\citep{li2022learning} combines heterogeneous knowledge, including phonetics, glyph and definition, in the dictionary through constrastive learning. SCOPE~\citep{li-etal-2022-improving-chinese} designs an auxiliary task called Chinese pronounciation prediction to improve model performance.

\subsection{Chinese Grammatical Error Correction}
CGEC can be treated as a Seq2Seq task~\citep{DBLP:journals/csur/DongLGCLSY23} like Neural Machine Translation (NMT). BLCU~\citep{ren2018sequence} employs a CNN-based Seq2Seq model to generate corrected sentence. AliGM~\citep{zhou2018chinese} utilizes a LSTM-based encoder-decoder model to resolve CGEC task. HRG~\citep{hinson2020heterogeneous} proposes an ensemble method containing a Seq2Seq model, a Seq2Tag model and a spelling checker.
Later CGEC works mainly use Transformer~\citep{vaswani2017attention} model which has achieved great success in NMT. TEA~\citep{wang2020chinese} and WCDA~\citep{tang2021chinese} employ Transformer encoder-decoder for CGEC task, and propose different data augmentation approaches by automatically corrupting input sentences. MaskGEC~\citep{zhao2020maskgec} adds dynamic random masks to the source sentences to enhance the generalization ability of Transformer-based CGEC model.
Recently, some works view CGEC as a Seq2Tag task, which predicts the edit operations for each token in the source sentence iteratively. \citet{liang2020bert} uses BERT-based sequence tagging model to detect and correct errors. \citet{zhang2022mucgec} adopts GECToR~\citep{omelianchuk2020gector} for CGEC task and enhances the model by applying Chinese PLMs. TtT~\citep{li2021tail} proposes a non-autoregressive model, which utilizes BERT as encoder and CRF~\citep{lafferty2001conditional} as decoder. S2A~\citep{li2022sequence} proposes a model that combines the advantages of Seq2Seq and Seq2Tag methods.

To the best of our knowledge, \methodName{} represents the first attempt to introduce the concept of progressive learning to the field of Chinese text error correction. Moreover, our framework is the first model-agnostic framework that can be deployed across various models for both text-aligned CSC and text-non-aligned CGEC tasks, thereby enhancing the error correction performance of these models.

\begin{figure*}[ht]
\centering
\includegraphics[width=0.95\textwidth]{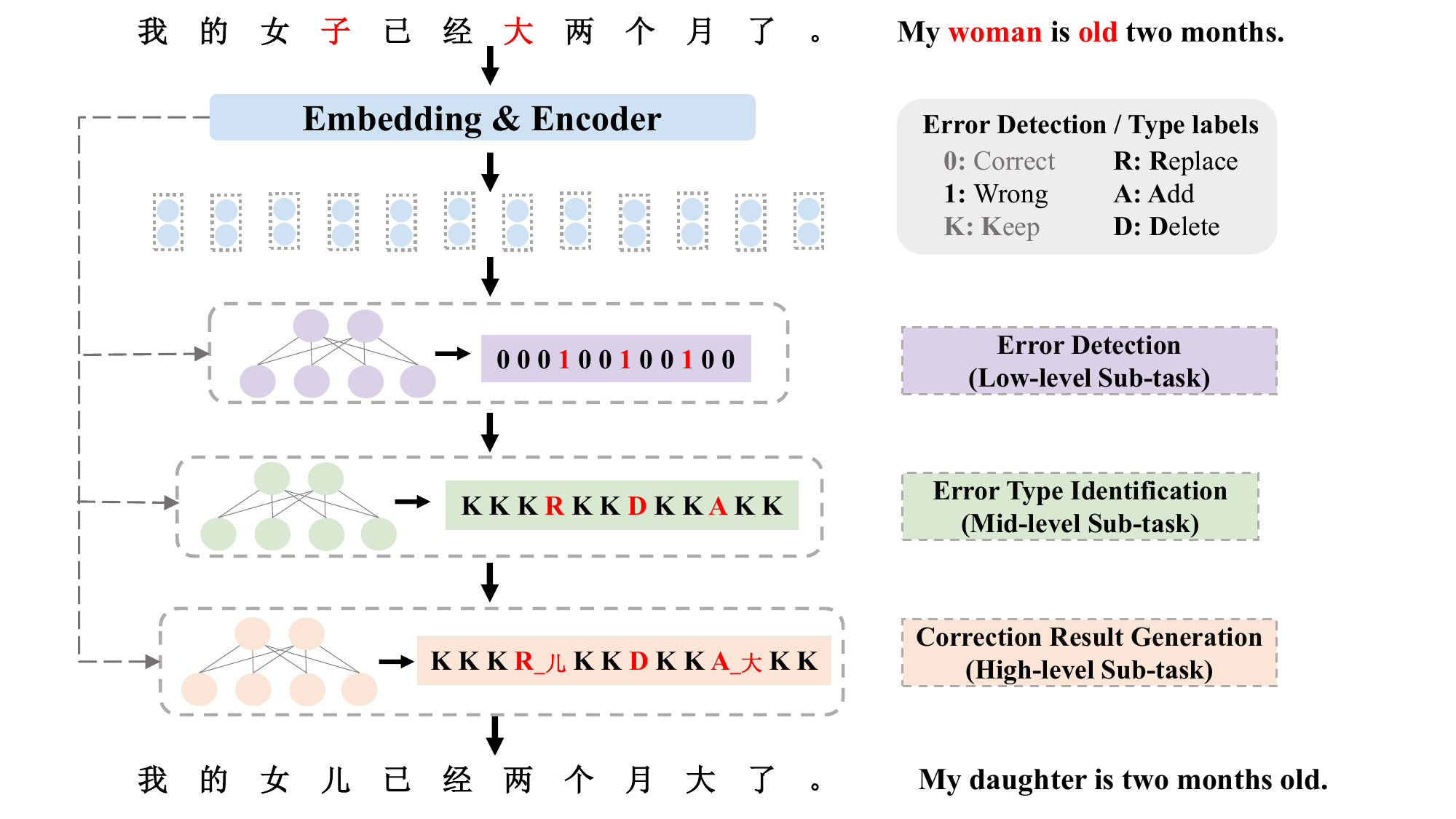}
\caption{Overview of our proposed \methodName{} framework, which aims to guide models to correct errors from easy to difficult, just like humans.}
\label{fig:overview}
\end{figure*}

\section{Methodology}
\subsection{Problem Formulation}
Chinese Text Error Correction (CTEC) task targets to detect and correct various types of errors in Chinese texts. 
Given an input sentence $X = \{x_1,x_2,...,x_n\}$ containing $m$ tokens, a CTEC model should correct the text errors in $X$, and output the corrected sentence $Y = \{y_1,y_2,...,y_m\}$. Specifically, the CSC subtask is a text-aligned task, thus the length of the output sentence is equal to that of the input sentence, or $m = n$.

The CTEC task can be naturally regarded as a Sequence-to-Sequence (Seq2Seq) conditional text generation task which models the conditional probability $p(Y|X;\theta)$ parameterized by $\theta$. Alternatively, CTEC can also be viewed as a Sequence-to-Tagging (Seq2Tag) task that needs to model $p(O|X;\theta)$, where $O=\{o_1, o_2,..., o_n\}$ is an edit operation sequence of the same length as $X$. The output sentence $Y$ can be obtained by applying the edit operations to each token in $X$, which can be formulated as $Y=\text{Concat}([o_1(x_1), o_2(x_2), ..., o_n(x_n)])$.

\subsection{Framework Overview}

The key motivation of our framework \methodName{} is to guide CTEC models to imitate the process of humans correcting text errors.
The core idea of our framework consists of two aspects. Firstly, we decompose the CTEC task into multiple sub-tasks, which allows the model to learn incrementally from simpler to more complex tasks, resulting in a more streamlined training process. Secondly, we enable the model to sequentially complete various sub-tasks, sequentially narrowing down the range of candidate results, thereby accomplishing the objective of reducing over-correction.

Specifically, as shown in Figure~\ref{fig:overview}, we divide the CTEC task into three sub-tasks from easy to difficult: Error Detection (Low-Level), Error Type Identification (Mid-Level), and Correction Result Generation (High-Level).

\textbf{Error Detection} sub-task targets to detect errors in a sentence. Given a sentence that may contain errors, the model should perform binary classification on each token in the sentence to determine whether it is an error or not.
\textbf{Error Type Identification} sub-task aims to analyze the error types of erroneous tokens in the sentence in order to narrow down the range of candidate results. For the tokens identified as errors in Error Detection sub-task, the model should further classifies them into predefined error types.
\textbf{Correction Result Generation} sub-task requires the model to generate the final correction result based on the outputs of previous sub-tasks.
During the training process, we employ these three sub-tasks to construct a multi-task training objective, guiding the model to learn the process of error correction progressively. During the inference process, the model would complete these sub-tasks sequentially, to achieve text error correction.

The commonly used CTEC models belong to the encoder-decoder architecture. The encoder part is typically composed of a pre-trained language model, which is employed to obtain the semantic representation of an input sentence. Meanwhile, the decoder part differs in different models. In a Seq2Tag model, the decoder part is usually a simple MLP that decodes the edit operation corresponding to each token in the sentence in parallel. In a Seq2Seq model, the decoder is typically a model of similar size to the encoder, which generates the corrected sentence autoregressively.

\methodName{} is a model-agnostic learning framework that imposes no restrictions on the specific forms of the encoder and decoder. It is applicable to various CTEC models with different architectures.

\subsection{Error Detection Sub-task (Low-Level)}
The goal of Error Detection Sub-task is to enable the CTEC model to determine whether each token in the sentence is erroneous.
Given an input sentence $X=\{x_1,x_2,...,x_n\}$, we employ the encoder of the model $E$ to encode it to obtain token-level semantic representations:
\begin{equation}
    H = E(X) = \{h_1,h_2,...,h_n\},
\end{equation}
where $h_i\in\mathbb{R}^d$ and $d$ is hidden size of the encoder model $E$.

Then, a simple MLP is utilized to compute the probability that each token is an erroneous token. The output probability of the $i$-th token in this sub-task can be represented as:
\begin{equation}
    p^1_i = \text{softmax}(W_{12}\cdot \sigma (W_{11}\cdot h_i + b_{11}) + b_{12}),
\end{equation}
where $p^1_i\in\mathbb{R}^{k_1}$, and $k_1$ is the size of label space. In this sub-task, $k_1 = 2$, which represents that the label of each token has two categories: correct and incorrect. Moreover, $\sigma(.)$ is the activation function, and $W_{11}\in\mathbb{R}^{d\times d}, W_{12}\in\mathbb{R}^{d\times k_1}, b_{11}\in\mathbb{R}^{d}, b_{12}\in\mathbb{R}^{k_1}$ are learnable parameters.

At this level, 
we use cross-entropy loss function which is widely used for classification tasks as the training objective:
\begin{equation}
    \mathcal{L}_1 = -\dfrac{1}{n}\sum^{n}_{i=1}{y_i^1 \log{p_i^1}},
\end{equation}
where $y_i^1$ is the ground truth label of the $i$-th token in this sub-task.
During inference, we select the index with a higher predicted probability as the predicted result for whether each token is erroneous.

\subsection{Error Type Identification Sub-task (Mid-Level)}
Based on the output obtained from the preceding Error Detection Sub-task (Low-Level), we can determine the correctness of each token within the sentence. Unaltered tokens correspond to correct instances, while erroneous tokens undergo subsequent processing in the Error Type Identification sub-task (Mid-Level). Error Type Identification sub-task targets to enable the model to analyze which category the erroneous tokens belong to predefined error types.

In the CGEC task, we define the error types of each token as types of edit operations, including Appending (A), Deleting (D), and Replacing (R). Meanwhile, in the CSC task, since only spelling errors are considered, we follow previous work and define the error types as two categories: Phonological Error (P) and Morphological error (M).

Similar to the previous sub-task, based on the token-level semantic representations encoded by the encoder, we employ a simple MLP to obtain the probability that each erroneous token belongs to each error category:
\begin{equation}
    p^2_i = \text{softmax}(W_{22}\cdot \sigma (W_{21}\cdot h_i + b_{21}) + b_{22}),
\end{equation}
where $p^2_i\in\mathbb{R}^{k_2}$, and $k_2$ is the size of label space. As mentioned above, $k_2 = 2$ for the CSC task and $k_2 = 3$ for the CGEC task. $W_{21}\in\mathbb{R}^{d\times d}, W_{22}\in\mathbb{R}^{d\times k_2}, b_{21}\in\mathbb{R}^{d}, b_{22}\in\mathbb{R}^{k_2}$ are also trainable parameters.

We also use the cross-entropy loss function as the training objective, but in this sub-task, only the loss values for erroneous tokens are calculated:
\begin{equation}
    \mathcal{L}_2 = -\dfrac{\displaystyle\sum^n_{i=1}{\vmathbb{1}(y^1_i = 1)\cdot y^2_i \log{p^2_i}}}{\displaystyle\sum^n_{i=1}{\vmathbb{1}(y^1_i = 1)}},
\end{equation}
where $y^2_i, y^1_i$ are the training labels of the $i$-th token in this sub-task and in the previous sub-task respectively, and $\vmathbb{1}(.)$ is the indicator function.
During inference, we also select the index with the highest predicted probability as the predicted error category for each erroneous token.

\subsection{Correction Result Generation Sub-task (High-Level)}
Similarly, the Correction Result Generation Sub-task aims to generate the final error correction result based on the output of the previous Error Type Identification Sub-task (Mid-Level), which is also the target of the CTEC task. The results of the Error Detection Sub-task reveal which tokens in the sentence are incorrect. For correct tokens, we would keep the original tokens unchanged. Meanwhile, for incorrect tokens, we would use the error types identified in the Error Type Identification Sub-task to narrow down the range of candidates for generation.

In particular, the error correction result generated by CSC models is an error-free sentence that is the same length as the input sentence. For correct tokens, the corresponding tokens are copied into the correction result. For tokens that are predicted to be phonological or morphological errors, we would select candidate tokens from phonological or morphological confusion set related to input tokens.
CGEC models can be divided into two architectures: Seq2Taq and Seq2Seq. A Seq2Tag model generates edit operations as correction results corresponding to input tokens, such as keeping, deleting, replacing with a specific token, appending a specific token, etc. Meanwhile, a Seq2Seq model simply generates the corrected sentence as the result.

In this sub-task, the token-level semantic representations output by the model's encoder are fed to the model's decoder, which generates the final error correction results. This step is similar to that of a normal model. For CSC models and Seq2Tag models in CGEC tasks, their decoders are typically simple linear transformations. For each token, the probability of correction results can be computed as
\begin{equation}
    p^3_i = \text{softmax}(W_3 h_i + b_3),
\end{equation}
where $p^3_i\in\mathbb{R}^v$, and $v$ is the size of the model's output vocabulary. $W_3\in\mathbb{R}^{d\times v}, b_3\in\mathbb{R}^{v}$ are trainable parameters.

Moreover, our framework employs the results of previous sub-tasks to constrain the search space of correction results. Specifically, we construct a mask matrix $M\in\mathbb{R}^{n\times v}$ with the same dimension as the output probability to represent the restrictions on candidate results:
\begin{equation}
m_{ij}=
\begin{cases}
1, & \text{if } V_j \in S(y^1_i, y^2_i, x_i, V), \\
0, & \text{otherwise,}
\end{cases}
\end{equation}
where $m_{ij}$ represents the element in the $i$-th row and $j$-th column of matrix $M$, indicating whether the candidate result for the $i$-th token in the output sequence includes the $j$-th token in the vocabulary. Moreover, $S(y^1_i, y^2_i, x_i, V)$ denotes a subset of the vocabulary $V$.

Then we perform element-wise multiplication between the output probability and the mask vector for restriction, resulting in the final probability of each candidate:
\begin{equation}
    p_i = p^3_i \odot m_i.
\end{equation}

\begin{figure*}[ht]
\centering
\includegraphics[width=0.95\textwidth]{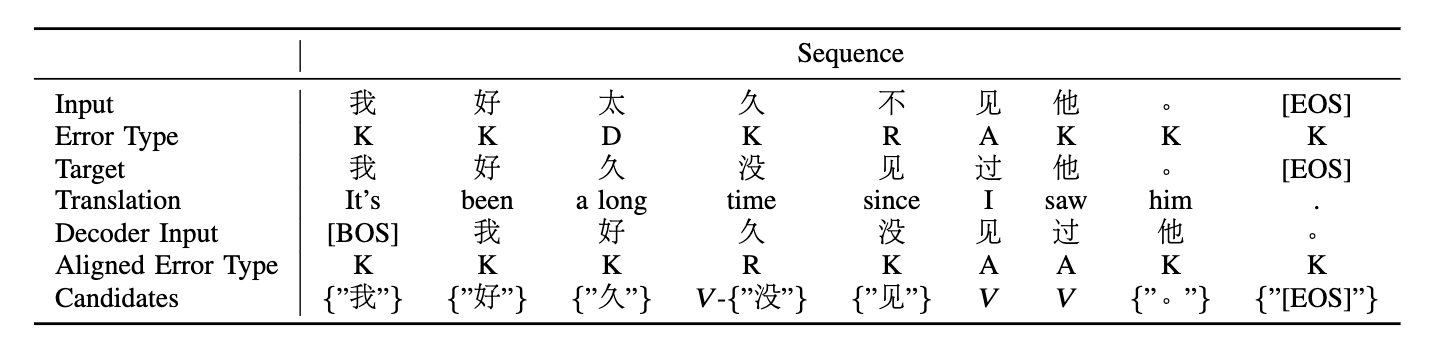}
\caption{An example of aligning error type labels corresponding to the input sequence with the output sequence. The error types contain 4 categories: \textbf{K}eep, \textbf{R}eplace, \textbf{A}ppend, and \textbf{D}elete. For candidates, $V$ refers to the entire vocabulary of the model.}
\label{tab:alignment}
\end{figure*}


For Seq2Seq models, the decoder is an autoregressive generation model that directly generates corrected text sequences. Consequently, the lengths of input and output sequences are often inconsistent. To address the issue, we extend the error type labels corresponding to the input sentence to the same length as the output sentence during training, based on certain rules. 
During inference, a state variable is maintained to indicate the correspondence between the output index and the input index, allowing the model to utilize the error type information output by the previous sub-task at the decoder.
Figure~\ref{tab:alignment} illustrates an example of aligning error type labels corresponding to the input sequence with the output sequence.

We also utilize cross-entropy as the loss function in this sub-task:
\begin{equation}
    \mathcal{L}_3 = -\dfrac{1}{n}\sum^n_{i=1}{y_i \log{p_i}},
\end{equation}
where $y_i$ is the ground truth of the $i$-th token in the CTEC task, i.e., the correct token or the correct edit operation corresponding to the input token.

\subsection{Training and Inference}
During the model training process, we construct the overall loss function as a weighted sum of the individual loss functions for the three sub-tasks described above:
\begin{equation}
    \mathcal{L} = \mathcal{L}_3 + \lambda \mathcal{L}_1 + \mu \mathcal{L}_2,
\end{equation}
where $\lambda$ and $\mu$ denote the weights of loss functions for Error Detection and Error Type Identification sub-tasks, respectively.
It is worth noting that, we expect the model to progressively learn each sub-task of text correction from easy to difficult during training, so the three sub-tasks are trained independently, i.e., it is the ground truth label for the $i$-th ($i=1,2$) sub-task instead of the model output that is employed as input for the $i+1$-th sub-task. In this way, the error propagation is reduced and the correction performance is enhanced.

During the inference process, we utilized the CTEC model to sequentially complete the three sub-tasks to generate the final correction results. Notably, for the Seq2Tag models in the CGEC task, it is possible for the output sentence to still contain errors after one round of editing, which necessitates multiple rounds of iterative editing.

\section{Experiments}
As mentioned in Section~\ref{sec:rw}, in the NLP research community, Chinese Text Error Correction (CTEC) consists of two tasks: Chinese Spelling Correction (CSC) and Chinese Grammatical Error Correction (CGEC). To verify the generality of our proposed framework, we conduct experiments on multiple text error correction models with different architectures for both tasks, and evaluate the model performance on multiple widely used datasets.

\begin{table}[ht]
\centering
\caption{Statistics of datasets used in our experiments.}
\label{tab:dataset}
\begin{tabular}{ccrr}
\toprule
\multicolumn{1}{l}{} & \textbf{Dataset} & \multicolumn{1}{c}{\textbf{\#Sent.}} & \multicolumn{1}{c}{\textbf{Length}} \\ 
\midrule
\multirow{4}{*}{\begin{tabular}[c]{@{}c@{}}CSC\\Training\end{tabular}} & SIGHAN13 & 700 & 41.8 \\
 & SIGHAN14 & 3,437 & 49.6 \\
 & SIGHAN15 & 2,338 & 31.3 \\
 & Wang271K & 271,329 & 42.6 \\ 
\midrule
\multirow{3}{*}{\begin{tabular}[c]{@{}c@{}}CSC\\Test\end{tabular}} & SIGHAN13 & 1,000 & 74.3 \\
 & SIGHAN14 & 1,062 & 50.0 \\
 & SIGHAN15 & 1,100 & 30.6 \\ 
\midrule
\multirow{2}{*}{\begin{tabular}[c]{@{}c@{}}CGEC\\Training\end{tabular}} & Lang8 & 1,220,906 & 18.9 \\
 & HSK & 156,870 & 27.3 \\ 
\midrule
\multirow{2}{*}{\begin{tabular}[c]{@{}c@{}}CGEC\\Test\end{tabular}} & NLPCC & 2,000 & 29.7 \\
 & MuCGEC (Dev.) & 1,137 & 44.0 \\
\bottomrule
\end{tabular}
\end{table}

\subsection{Datasets}
Following previous CSC and CGEC work, We employ widely used datasets for model training and evaluation.


For the CSC task, we use the test sets of SIGHAN13~\citep{wu2013chinese}, SIGHAN14~\citep{yu2014overview}, and SIGHAN15~\citep{tseng2015introduction} as the evaluation data. The training data consists of manually annotated samples from SIGHAN13/14/15, and 271K automatically generated samples (denoted as Wang271K)~\citep{wang2018hybrid}. Training and test setting is exactly the same as previous works~\citep{liu2021plome,xu2021read,li2022learning}.
The corpora of SIGHAN datasets are all collected from the essay section of the computer-based Test of Chinese as a Foreign Language (TOCFL), administrated in Taiwan. Note that the original sentences of SIGHAN datasets are in Traditional Chinese. Following previous works~\citep{liu2021plome,xu2021read,li2022learning}, we utilized OpenCC\footnote{https://github.com/BYVoid/OpenCC} toolkit to convert them to Simplified Chinese.


For the CGEC task, two evaluation benchmarks are employed to evaluate the model performance: NLPCC~\citep{zhao2018overview} and MuCGEC~\citep{zhang2022mucgec}.
Besides, Lang8~\citep{zhao2018overview} and HSK~\citep{zhang2009features} are used for model training.
For a fair comparison, the use of the above training set and verification set is the same as the previous works~\citep{zhao2020maskgec,tang2021chinese}. 
NLPCC is a shared task introduced in 2018, wherein the training data is collected from Lang8, a language learning website, while the test data is extracted from PKU Chinese Learner Corpus containing Chinese essays written by foreign college students. MuCGEC extracts a subset of sentences from different datasets, including NLPCC, Lang8, and CGED~\citep{rao2018overview, rao2020overview}, and re-annotated them to create a new high-quality evaluation dataset. 
Table~\ref{tab:dataset} presents the statistics of datasets used in our experiments.

\subsection{Evaluation Metrics}
For the CSC task, previous works~\citep{xu2021read, li2022past, li2022learning} employ classification metrics to evaluate the model performance. In terms of evaluation granularity, there are two levels of evaluation metrics, character-level metrics and sentence-level metrics. Since there may be more than one wrong characters in a sentence, sentence-level metrics are more challenging than character-level metrics. We report sentence-level metrics in our experiments.
Specifically, we report Precision, Recall and $\text{F}_1$ score for both detection level and correction level. At the detection level, the model is required to identify all erroneous positions in a sentence. At the correction level, the model should not only detect but also correct all the wrong characters to the ground truths.

For CGEC, we use the official evaluation metrics provided by each benchmark. Specifically, the MaxMatch ($\text{M}^2$) scorer is employed in the NLPCC benchmark, and the ChERRANT scorer is employed in MuCGEC. 
$\text{M}^2$ scorer computes the phrase-level edits between the source sentence and the model output, while ChERRANT scorer calculates character-based span-level edits, avoiding the dependence of $\text{M}^2$ scorer on Chinese word segmentation tools.
We report Precision, Recall, and $\text{F}_{0.5}$ calculated by these scorers. Following previous works~\citep{zhang2022mucgec, li2022sequence, zhao2018overview}, we adopt $\text{F}_{0.5}$ as the evaluation metric instead of $\text{F}_1$, as precision is considered more crucial than recall in the CGEC task.

\subsection{Compared Baselines}
To evaluate the performance of \methodName{}, we select some strong CSC and CGEC models as our baselines, including previous state-of-the-art methods. It is worth noting that although Large Language Models (LLMs) have achieved good performance on many NLP downstream tasks, existing studies~\citep{li2023effectiveness, DBLP:journals/corr/abs-2304-01746, DBLP:journals/corr/abs-2307-03972} have fully shown that LLMs have a large gap with the automatic evaluation metrics of CTEC due to the uncontrollability and freedom of the content they generate. So LLMs still cannot achieve performance that can compete with supervised training of small models on CTEC. Therefore, in this work, we do not choose LLMs as baselines.
For the CSC task, we compare the following models:
\begin{enumerate}
    \item \textbf{MacBERT}~\citep{cui2020revisiting} is an enhanced BERT~\citep{devlin2019bert} model which utilizes similar words instead of [MASK] symbol for MLM pre-training task. It is a stronger baseline than BERT in the CSC task.
    \item \textbf{SpellGCN}~\citep{cheng2020spellgcn} employs GCN to incorporate phonetic and visual knowledge and model the character similarity for the CSC task.
    \item \textbf{PLOME}~\citep{liu2021plome} introduces phonetics and stroke information by GRU encoder and pre-trains a task-oriented BERT model with a masking strategy based on the confusion set.
    \item \textbf{MLM-phonetics}~\citep{zhang2021correcting} employs ERNIE~\citep{sun2020ernie} model for CSC task and adopts phonetic confusion character replacement strategy to integrate additional phonetic information into the model.
    \item \textbf{REALISE}~\citep{xu2021read} is a multi-modal CSC model that uses multiple models to mix semantic, phonetic, and visual information of input characters.
    \item \textbf{LEAD}~\citep{li2022learning} leverages heterogeneous knowledge in the dictionary through a constrastive learning method.
    \item \textbf{SCOPE}~\citep{li-etal-2022-improving-chinese} introduces an auxiliary task named Chinese pronunciation prediction to improve the CSC performance, which is the current state-of-the-art method.
\end{enumerate}

For the CGEC task, the following baselines are chosen:
\begin{enumerate}
    \item \textbf{HRG}~\citep{hinson2020heterogeneous} is a heterogeneous approach composed of a Seq2Seq model, a Seq2Tag model and a spelling checker.
    \item \textbf{MaskGEC}~\citep{zhao2020maskgec} dynamically adds random masks to the source sentences during the training procedure to construct diverse pseudo-training pairs, which enhances the generalization ability of the CGEC model.
    \item \textbf{TEA}~\citep{wang2020chinese} proposes a dynamic residual structure to enhance the Transformer-based CGEC model, and employs a data augmentation method by corrupting input sentences for model training.
    \item \textbf{WCDA}~\citep{tang2021chinese} proposes a data augmentation method combining character and word granularity noise to improve the performance of the Transformer-based CGEC model.
    \item \textbf{S2A}~\citep{li2022sequence} proposes a Sequence-to-Action module which combines the advantages of both Seq2Seq and Seq2Tag approaches.
    \item \textbf{BART-Chinese}~\citep{shao2021cpt} is a Chinese version of BART~\citep{lewis2020bart}. BART is a Seq2Seq pre-training model based on Transformer architecture, which uses a denoising auto-encoder (DAE) as its pre-training objective.
    \item \textbf{GECToR-Chinese}~\citep{zhang2022mucgec} is a Chinese variant of GECToR~\citep{omelianchuk2020gector} which utilizes StructBERT~\citep{wei2020structbert} as backbone. GECToR is an iterative sequence tagging grammatical error correction approach.
\end{enumerate}

\subsection{Implementation Details}
We utilize PyTorch~\citep{paszke2019pytorch}, Huggingface's Transformers~\citep{wolf2020transformers} and Pytorch-Lightning to implement \methodName{}. For the CSC task, we employ \methodName{} on MacBERT and SCOPE models to improve their performance. For  GEC, we utilize our framework on BART-Chinese and GECToR-Chinese models. The loss weights of both Error Detection and Error Type Identification sub-tasks $\lambda$ and $\mu$ are set to $1.0$. 
Other hyper-parameters utilized for training and inference are set in accordance with the corresponding original papers.

All of our experiments are run on a Linux server equipped with 64 Intel(R) Xeon(R) Gold 6326 CPU cores and 8 NVIDIA GeForce RTX 3090 GPUs, and each experiment takes up 1 or 2 GPUs.

\begin{table}[ht]
\centering
\tiny
\caption{The performance of \methodName{} and all baselines. \methodName{} (X) means that we combine \methodName{} with model X.}
\label{tab:main_results_csc}
\begin{tabular}{c|l|ccc|ccc} 
\toprule
\multirow{2}{*}{\textbf{Dataset}} & \multicolumn{1}{c|}{\multirow{2}{*}{\textbf{Method}}} & \multicolumn{3}{c|}{\textbf{Detection Level}} & \multicolumn{3}{c}{\textbf{Correction Level}} \\
 & \multicolumn{1}{c|}{} & \textbf{Pre} & \textbf{Rec} & $\textbf{F}_1$ & \textbf{Pre} & \textbf{Rec} & $\textbf{F}_1$ \\ 
\cmidrule(lr){1-8}
\multirow{8}{*}{\textbf{SIGHAN13}} & SpellGCN~\citep{cheng2020spellgcn} & 80.1 & 74.4 & 77.2 & 78.3 & 72.7 & 75.4 \\
 & MLM-phonetics~\citep{zhang2021correcting} & 82.0 & 78.3 & 80.1 & 79.5 & 77.0 & 78.2 \\
 & REALISE~\citep{xu2021read} & \textbf{88.6} & 82.5 & 85.4 & 87.2 & 81.2 & 84.1 \\
 & LEAD~\citep{li2022learning} & 88.3 & 83.4 & 85.8 & 87.2 & 82.4 & 84.7 \\ 
\cmidrule(lr){2-8}
 & MacBERT~\citep{cui2020revisiting} & 84.6 & 79.9 & 82.2 & 81.0 & 76.5 & 78.7 \\
 & \methodName{} (MacBERT) & $87.8$ & $81.6$ & $84.6$ & $86.7$ & $80.5$ & $83.5$ \\ 
\cmidrule(lr){2-8}
 & SCOPE~\citep{li-etal-2022-improving-chinese} & 87.4 & 83.4 & 85.4 & 86.3 & 82.4 & 84.3 \\
 & \methodName{} (SCOPE) & $88.5$ & $\textbf{83.7}$ & $\textbf{86.0}$ & $\textbf{87.7}$ & $\textbf{83.0}$ & $\textbf{85.3}$ \\ 
\midrule
\multirow{8}{*}{\textbf{SIGHAN14}} & SpellGCN~\citep{cheng2020spellgcn} & 65.1 & 69.5 & 67.2 & 63.1 & 67.2 & 65.3 \\
 & REALISE~\citep{xu2021read} & 67.8 & 71.5 & 69.6 & 66.3 & 70.0 & 68.1 \\
 & MLM-phonetics~\citep{zhang2021correcting} & 66.2 & \textbf{73.8} & 69.8 & 64.2 & \textbf{73.8} & 68.7 \\
 & LEAD~\citep{li2022learning} & \textbf{70.7} & 71.0 & 70.8 & \textbf{69.3} & 69.6 & 69.5 \\ 
\cmidrule(lr){2-8}
 & MacBERT~\citep{cui2020revisiting} & 63.9 & 65.6 & 64.7 & 61.0 & 62.7 & 61.9 \\
 & \methodName{} (MacBERT) & $65.6$ & $68.8$ & $67.2$ & $64.8$ & $68.1$ & $66.4$ \\ 
\cmidrule(lr){2-8}
 & SCOPE~\citep{li-etal-2022-improving-chinese} & 70.1 & 73.1 & 71.6 & 68.6 & 71.5 & 70.1 \\
 & \methodName{} (SCOPE) & $70.2$ & $73.3$ & $\textbf{71.7}$ & $\textbf{69.3}$ & $72.3$ & $\textbf{70.7}$ \\ 
\midrule
\multirow{8}{*}{\textbf{SIGHAN15}} & SpellGCN~\citep{cheng2020spellgcn} & 75.6 & 80.4 & 77.9 & 73.2 & 77.8 & 75.4 \\
 & PLOME~\citep{liu2021plome} & 77.4 & 81.5 & 79.4 & 75.3 & 79.3 & 77.2 \\
 & MLM-phonetics~\citep{zhang2021correcting} & 77.5 & 83.1 & 80.2 & 74.9 & 80.2 & 77.5 \\
 & REALISE~\citep{xu2021read} & 77.3 & 81.3 & 79.3 & 75.9 & 79.9 & 77.8 \\
 & LEAD~\citep{li2022learning} & 79.2 & 82.8 & 80.9 & 77.6 & 81.2 & 79.3 \\ 
\cmidrule(lr){2-8}
 & MacBERT~\citep{cui2020revisiting} & 71.9 & 77.9 & 74.8 & 68.0 & 73.6 & 70.7 \\
 & \methodName{} (MacBERT) & 71.1 & $80.2 $ & $75.4$ & $69.3$ & $78.2$ & $73.5$ \\ 
\cmidrule(lr){2-8}
 & SCOPE~\citep{li-etal-2022-improving-chinese} & 81.1 & 84.3 & 82.7 & 79.2 & \textbf{82.3} & 80.7 \\
 & \methodName{} (SCOPE) & $\textbf{82.9}$ & $\textbf{84.8}$ & $\textbf{83.8}$ & $\textbf{80.3}$ & $\textbf{82.3} $ & $\textbf{81.3}$ \\
\bottomrule
\end{tabular}
\end{table}

\subsection{Overall Performance}

The main experimental findings are detailed in Table~\ref{tab:main_results_csc} and Table~\ref{tab:main_results_cgec}. Table~\ref{tab:main_results_csc} presents the experimental outcomes concerning the SIGHAN13/14/15 datasets within the context of the CSC task, while Table~\ref{tab:main_results_cgec} delineates the experimental results for NLPCC and MuCGEC in the CGEC task. 
The analysis reveals that subsequent to training diverse models with \methodName{}, notable performance enhancements are observed across multiple datasets corresponding to the CTEC tasks. Specifically, both MacBERT and SCOPE models in the CSC task, as well as BART-Chinese and GECToR-Chinese models in the CGEC task, exhibit substantial improvements.

\begin{table}[ht]
\centering
\small
\caption{The performance of \methodName{} and all baselines. \methodName{} (X) means that we combine \methodName{} with model X.}
\label{tab:main_results_cgec}
\resizebox{\linewidth}{!}{
\begin{tabular}{l|ccc|ccc} 
\toprule
 & \multicolumn{3}{c|}{\textbf{NLPCC}} & \multicolumn{3}{c}{\textbf{MuCGEC}} \\
 & \textbf{Pre} & \textbf{Rec} & $\textbf{F}_{0.5}$ & \textbf{Pre} & \textbf{Rec} & $\textbf{F}_{0.5}$ \\ 
\midrule
HRG~\citep{hinson2020heterogeneous} & 36.79 & 27.82 & 34.56 & - & - & - \\
MaskGEC~\citep{zhao2020maskgec} & 44.36 & 22.18 & 36.97 & 35.95 & 21.72 & 31.60 \\
TEA~\citep{wang2020chinese} & 39.43 & ~22.80 & 34.41 & - & - & - \\
WCDA~\citep{tang2021chinese} & \textbf{47.41} & 23.72 & 39.51 & 40.11 & 22.87 & 34.86 \\
S2A~\citep{li2022sequence} & 42.34 & 27.11 & 38.06 & - & - & - \\ 
\midrule
BART-Chinese~\citep{shao2021cpt} & 41.44 & \textbf{32.89} & 39.39 & 37.93 & 28.46 & 35.56 \\
\methodName{} (BART) & $\text{46.57}$ & 31.46 & $\textbf{42.49}$ & $\text{40.87}$ & 28.30 & $\textbf{37.54}$ \\ 
\midrule
GECToR-Chinese~\citep{zhang2022mucgec}  & 42.88 & 30.19 & 39.55 & 38.88 & 27.95 & 36.06 \\
\methodName{} (GECToR) & $\text{45.90}$ & 29.10 & $\text{41.15}$ & $\textbf{41.32}$ & 27.25 & $\text{37.45}$ \\
\bottomrule
\end{tabular}
}
\end{table}

On one hand, the results indicate that the approach of devising multiple easy-to-difficult sub-tasks for CTEC models to learn progressively in our framework is effective in allowing the models to better understand and complete the text error correction task.
On the other hand, the results demonstrate that our framework is indeed model-agnostic and can be applied to various CTEC models of different architectures with high generality. In particular, one of the great advantages of \methodName{} is that it can be applied to both text-aligned CSC  and text-non-aligned CGEC tasks, thereby achieving unified modeling of the CTEC field.

\subsection{Efficiency Analysis}
We conduct experiments to compare the time consumed by model training and inference with and without \methodName{} using the same hyper-parameter settings and hardware configuration.

\begin{table}[ht]
\centering
\small
\caption{Training and inference time of CSC and CGEC models in a single epoch, both with and without our framework.}
\label{tab:efficiency}
\begin{tabular}{c!{\vrule width \lightrulewidth}cc|cc} 
\toprule
\multirow{2}{*}{\textbf{Method}} & \multicolumn{2}{c|}{\textbf{w/o \methodName{}}} & \multicolumn{2}{c}{\textbf{\textbf{w/ \methodName{}}}} \\
 & \textbf{Training} & \textbf{Inference} & \textbf{Training} & \textbf{Inference} \\ 
\midrule
MacBERT & $\text{1138}\pm\text{24s}$ & $\text{21}\pm\text{2s}$ & $\text{1235}\pm\text{26s}$ & $\text{23}\pm\text{3s}$ \\
SCOPE & $\text{1813}\pm\text{34s}$ & $\text{32}\pm\text{3s}$ & $\text{1987}\pm\text{39s}$ & $\text{34}\pm\text{3s}$ \\ 
\midrule
GECToR & $\text{8854}\pm\text{107s}$ & $\text{25}\pm\text{2s}$ & $\text{9538}\pm\text{101s}$ & $\text{26}\pm\text{3s}$ \\
BART & $\text{14187}\pm\text{128s}$ & $\text{368}\pm\text{24s}$ & $\text{15656}\pm\text{202s}$ & $\text{383}\pm\text{27s}$ \\
\bottomrule
\end{tabular}
\end{table}

Experimental results presented in Table~\ref{tab:efficiency} reveal that the training time increases by 5\%-10\% on average upon the adoption of \methodName{}, while the inference time experiences a modest increase of approximately 5\%. Indeed, for per individual sentence of the dataset, the average incremental cost of \methodName{} is inconsequential and the incurred time overhead falls within an acceptable range, demonstrating the high efficiency of our framework.

Compared to the original CTEC model, the additional time overhead incurred by \methodName{} mainly stems from three sub-tasks. Through a qualitative analysis, we identify two reasons for the high efficiency of our framework:
\begin{enumerate}
    \item Although \methodName{} requires the model to perform three sub-tasks sequentially during both training and inference, each sub-task can be highly parallelized, thus minimizing the impact on the overall temporal efficiency.
    \item The number of parameters added by our framework to the optimized CTEC model is insignificant compared to the number of parameters of the pre-trained language model itself, rendering the computational cost of the model executing various sub-tasks modest.
\end{enumerate}

\subsection{Ablation Studies}

To thoroughly evaluate the efficacy of each sub-task within our progressive learning framework, we conduct two groups of ablation experiments. These experiments involved examining the performance outcomes resulting from the removal of modules and conducting upper-bound analysis. The aim was to validate the effectiveness of the sub-tasks outlined in our framework.

\begin{table}[ht]
\centering
\small
\caption{Results of ablation experiments on SIGHAN14 dataset.}
\label{tab:remove_CSC}
\begin{tabular}{l|cccccc}
\toprule
 & \multicolumn{3}{c}{\begin{tabular}[c]{@{}c@{}}\textbf{Detection Level}\\\end{tabular}} & \multicolumn{3}{c}{\textbf{Correction Level}} \\
 & \textbf{Pre} & \textbf{Rec} & $\textbf{F}_1$ & \textbf{Pre} & \textbf{Rec} & $\textbf{F}_1$ \\ 
\midrule
\methodName{} (MacBERT) & 65.6 & 68.8 & 67.2 & 64.8 & 68.1 & 66.4 \\
~w/o ED & 64.9 & 67.1 & 66.0 & 62.7 & 64.8 & 63.7 \\
~w/o ETI & 64.8 & 66.9 & 65.8 & 62.5 & 64.3 & 63.4 \\
MacBERT & 63.9 & 65.5 & 64.7 & 61.0 & 62.7 & 61.9 \\
\bottomrule
\end{tabular}
\end{table}

\begin{table}[ht]
\centering
\small
\caption{Results of ablation experiments on NLPCC dataset.}
\label{tab:remove_CGEC}
\begin{tabular}{l|ccc} 
\toprule
 & \textbf{Pre} & \textbf{Rec} & $\textbf{F}_{0.5}$ \\ 
\midrule
\methodName{} (BART) & 46.57 & 31.46 & 42.49 \\
~w/o ED & 44.79 & 32.02 & 41.48 \\
~w/o ETI & 44.32 & 32.25 & 41.23 \\
BART-Chinese & 41.44 & 32.89 & 39.39  \\
\bottomrule
\end{tabular}
\end{table}

\subsubsection{Module Performance}
As evident from the results presented in Table~\ref{tab:remove_CSC} and Table~\ref{tab:remove_CGEC}, we conduct ablation experiments in which we remove the Error Detection (ED) and Error Type Identification (ETI) sub-tasks from \methodName{}. We then analyze the impact of these removals on the model's performance. Specifically, by removing the Error Detection (ED) sub-task, the model first identifies which error types of tokens (e.g., Delete or Replace) and then corrects them. On the other hand, removing the Error Type Identification (ETI) sub-task means that the model first identifies which tokens are wrong and then directly corrects them. It is important to note that the removal of both sub-tasks causes the model to revert to baseline performance.
By analyzing the outcomes illustrated in Table~\ref{tab:remove_CSC} and Table~\ref{tab:remove_CGEC}, we conduct an extensive analysis of two distinct datasets: the SIGHAN14 dataset for the CSC task and the NLPCC dataset for the CGEC task. 
The experimental results clearly demonstrate that the model's performance suffers a decline when either the Error Detection (ED) or the Error Type Identification (ETI) sub-task is removed. 
This demonstrates that the model's proficiency in identifying error tokens and discerning their respective error types is compromised when the error detection (ED) subtask is disregarded. Simultaneously, when the error type identification (ETI) subtask is omitted, the model's capacity to determine the error type of erroneous tokens while effecting corrections is also inadequate. It is notable that even in the absence of one sub-task, our model demonstrates superior performance compared to a single-task approach. These findings emphasize the effectiveness of these sub-tasks within \methodName{}, as they aid the model in mitigating error propagation through the paradigm of progressive learning.

\begin{table}[ht]
\centering
\small
\caption{Model performance when given ground truth labels for low-level and mid-level sub-tasks on SIGHAN15 dataset.}
\label{tab:upper_CSC}
\begin{tabular}{l|cccccc} 
\toprule
 & \multicolumn{3}{c}{\textbf{Detection Level}} & \multicolumn{3}{c}{\textbf{\textbf{Correction Level}}} \\
 & \textbf{Pre} & \textbf{Rec} & \multicolumn{1}{l}{$\textbf{F}_1$} & \textbf{Pre} & \textbf{Rec} & $\textbf{F}_1$ \\ 
\midrule
SCOPE & 81.1 & 84.3 & 82.7 & 79.2 & 82.3 & 80.7 \\
\methodName{} (SCOPE) & 82.9 & 84.8 & 83.8 & 80.3 & 82.3 & 81.3 \\
~w/ Low Label & 89.8 & 94.1 & 91.9 & 85.7 & 89.8 & 87.7 \\
~w/ Low \& Mid Label & 90.2 & 95.2 & 92.6 & 87.0 & 91.9 & 89.4 \\
\bottomrule
\end{tabular}
\end{table}

\subsubsection{Upper-Bound Analysis}
To assess the effectiveness and significance of our designed sub-tasks, we further conduct an upper bound analysis within the SIGHAN15 dataset of the CSC task, as delineated in Table~\ref{tab:upper_CSC}. Throughout the inference process, we substitute the predicted outcomes of both the Error Detection and Error Type Identification sub-tasks with the ground-truth labels. In other words, we use the golden answers from these two sub-tasks as additional input information for models and observe the changes in model performance. Utilizing the responses from the Error Detection and Error Type Identification sub-tasks as additional input information notably enhances the model's performance in error correction. This underscores the notion that guiding the model through a progression of sub-tasks, from simple to complex, contributes to an optimal resolution of the CTEC task. Furthermore, it underscores the rationale and efficacy of our approach, as demonstrated by the observed improvements.

\subsection{Analysis of Various Sub-tasks}

\begin{table}[ht]
\centering
\small
\caption{The \methodName{} (SCOPE) Model's performance of each sub-task on SIGHAN15.}
\label{tab:subtask1}
\begin{tabular}{cccc}
\toprule
\textbf{Sub-task} & \textbf{Precision} & \textbf{Recall} & $\textbf{F}_1$ \\ 
\midrule
Error Detection & 82.0 & 85.5 & 83.7 \\
Error Type Identification & 80.5 & 82.8 & 81.6 \\
Correction Result Generation & 80.3 & 82.3 & 81.3 \\
\bottomrule
\end{tabular}
\end{table}

To further analyze the different levels of our proposed sub-tasks, we report the performance of \methodName{} (SCOPE) model on each sub-task on the SIGHAN15 dataset. 
Experimental results in Table~\ref{tab:subtask1} reveal a gradual decrease in $\text{F}_1$ scores across the three sub-tasks, indicating that the difficulty of these sub-tasks progressively increases, with subsequent tasks presenting more challenges than their predecessors.
Additionally, we observe that the performance gap between different sub-tasks is not significant, reflecting that the model prediction results from earlier sub-tasks effectively guide the model in accomplishing subsequent sub-tasks without causing severe error propagation. Therefore, the common error propagation phenomenon in the progressive learning paradigm is not obvious in our proposed \methodName{} and does not affect the model performance.

\begin{figure}[h]
\centering
\setlength{\abovecaptionskip}{0.5em}
\subfloat[Effect of $\lambda$ on $\text{F}_{\text{0.5}}$] 
{ 
\includegraphics[width=0.46\columnwidth]{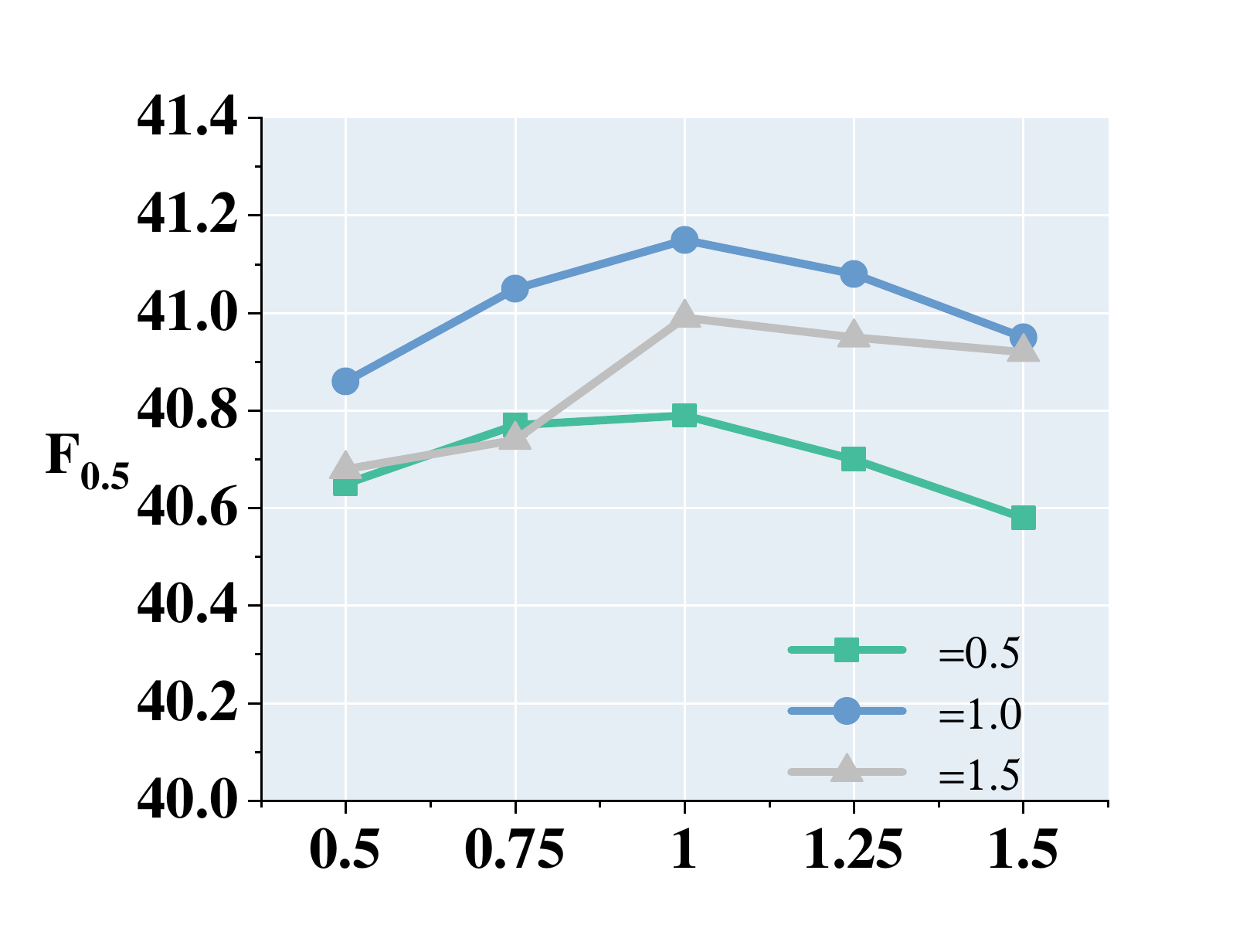} 
}
\subfloat[Effect of $\mu$ on $\text{F}_{\text{0.5}}$] 
{
\includegraphics[width=0.46\columnwidth]{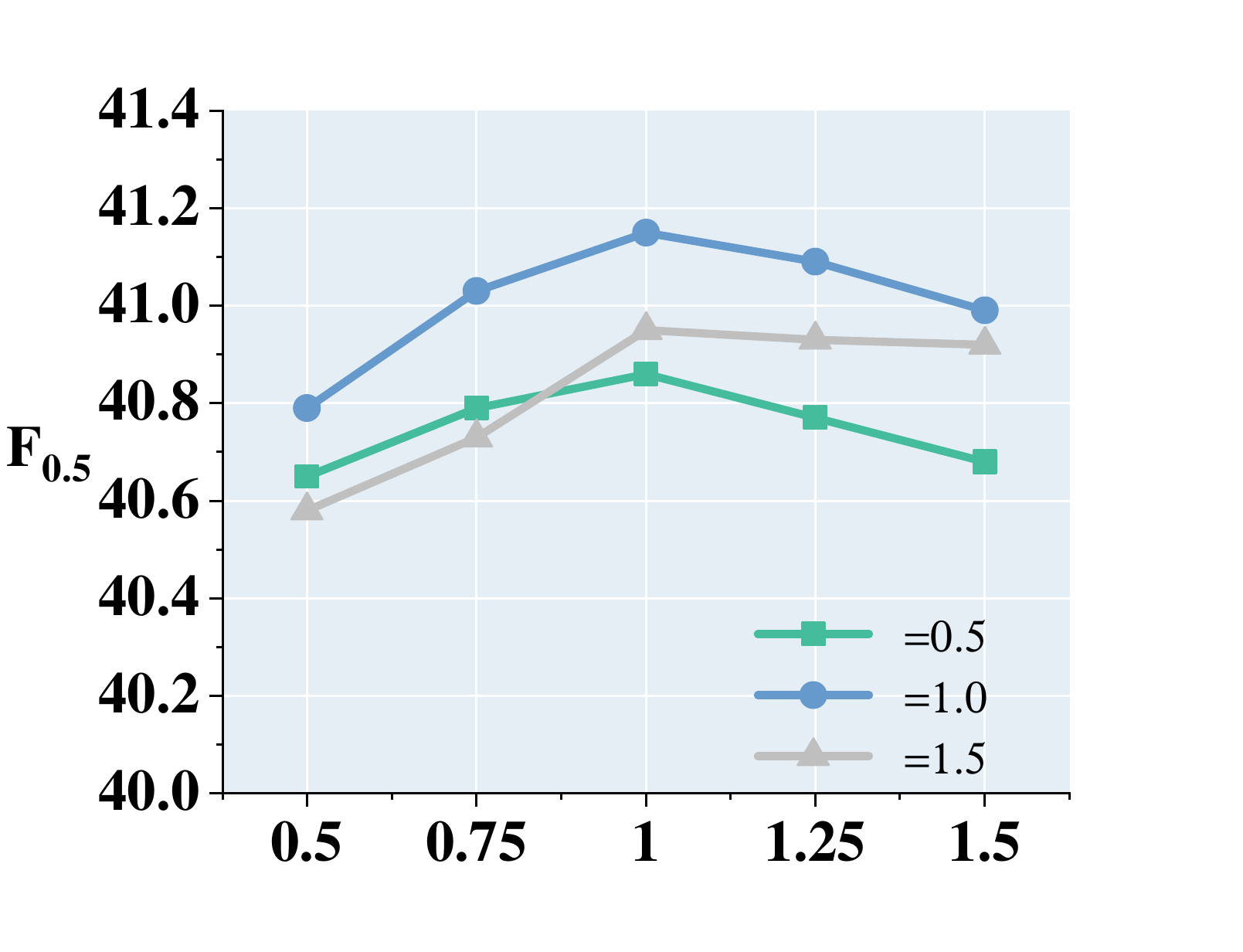} 
} 

\caption{Hyper-Parameter sensitivity analysis of $\lambda$ and $\mu$ on NLPCC dataset. We report the effect of varying one hyper-parameter while fixing the other on the $\text{F}_{\text{0.5}}$ metric.} 
\label{fig:sensitivity} 
\end{figure} 

\subsection{Hyper-parameter Sensitivity Analysis}
There is an important set of hyper-parameters in \methodName{}: the loss weights for Error Detection and Error Type Identification sub-tasks, $\lambda$ and $\mu$, during the training process. We adjust the loss weights to analyze how these hyper-parameters affect the performance of CTEC models.

From Figure~\ref{fig:sensitivity},
we can observe that:
\begin{enumerate}
    \item When the loss weights $\lambda$ and $\mu$ are similar and close to $1.0$ (i.e. the same as the weight for the Correction Result Generation sub-task), the CTEC models can achieve better performance. Therefore, we select $\lambda=\mu=1.0$ as the hyper-parameters for our main experiments.
    \item When the loss weights for the two sub-tasks are around $1.0$, the effect of changing them on model performance is relatively small, which reveals that the model performance is not sensitive to these hyper-parameters in our framework.
\end{enumerate}

\subsection{Case Study}
\begin{CJK*}{UTF8}{gbsn}

\begin{table}[ht]
\centering
\small
\caption{Two cases from MuCGEC evaluation dataset.}
\label{tab:case_study}
\resizebox{\columnwidth}{!}{%
\begin{tabular}{ll} 
\toprule
\multicolumn{2}{c}{\textbf{Case 1}} \\ 
\midrule
Input: & 我喜欢吃栗子甚于剥栗子。 \\
 & I like eating chestnuts more than peeling them. \\ 
\midrule
Reference: & 我喜欢吃栗子甚于剥栗子。 \\ 
\midrule
GECToR-Chinese: & 我喜欢吃栗子\textcolor{red}{，多}剥栗子。 \\
 & I like eating chestnuts, so peel them more. \\ 
\midrule
\methodName{} (GECToR): & 我喜欢吃栗子甚于剥栗子。 \\ 
\midrule
\multicolumn{2}{c}{\textbf{Case 2}} \\ 
\midrule
Input: & 吸烟不但损害身体健康，它也会\textcolor{red}{使人}引起疾病。 \\
 & Smoking not only damages health but also causes diseases. \\ 
\midrule
Reference: & 吸烟不但损害身体健康，它也会引起疾病。 \\ 
\midrule
BART-Chinese: & 吸烟不但损害身体健康，\textcolor[rgb]{1,0.502,0}{而且}也会引起疾病。 \\ 
\midrule
\methodName{} (BART): & 吸烟不但损害身体健康，它也会引起疾病。 \\
\bottomrule
\end{tabular}
}
\end{table}

Table~\ref{tab:case_study} presents two cases from MuCGEC dataset. In Case 1, we can observe that the baseline model BART-Chinese modifies the input sentence ``我喜欢吃栗子甚于剥栗子 (I like eating chestnuts more than peeling them)'' to ``我喜欢吃栗子，多剥栗子 (I like eating chestnuts, so peel them more)''. Although the modified sentence is grammatically correct, it erroneously alters the source sentence's meaning. In contrast, \methodName{} (BART) successfully avoids this unnecessary modification.
Similarly, in Case 2, the baseline model GECToR-Chinese correctly removes the redundant ``使人'' from the input sentence but superfluously changes ``它'' to ``而且''. Meanwhile, \methodName{} (GECToR) not only corrects the grammatical error but also better preserves the original expression of the sentence.
These cases further validate the effectiveness of our proposed progressive learning framework for CTEC, which enhances the performance of CTEC models while mitigating the issue of over-correction by mimicking the human error correction process.

\end{CJK*}
\section{Conclusion}
In this paper, we propose a model-agnostic progressive learning framework, named \methodName{}, for the Chinese Text Error Correction (CTEC) task. Our framework comprises three sub-tasks, arranged in an increasing order of difficulty: Error Detection, Error Type Identification, and Correction Result Generation. During training, the model progressively learns these sub-tasks, while during inference, it performs these sub-tasks sequentially to narrow down the candidates and generate the final correction results.
Extensive experiments indicate that \methodName{} can effectively enhance the correction performance of models with different architectures, further demonstrating the efficiency and effectiveness of our framework.
In the future, we intend to design more sophisticated sub-tasks to fully exploit the information in CTEC task, and investigate automatic construction of progressive sub-tasks to further enhance the generality and scalability of our framework.

\section*{Acknowledgments}
This research is supported by National Natural Science Foundation of China (Grant No.62276154), Research Center for Computer Network (Shenzhen) Ministry of Education, Beijing Academy of Artificial Intelligence (BAAI), the Natural Science Foundation of Guangdong Province (Grant No. 2023A1515 \\ 012914), Basic Research Fund of Shenzhen City (Grant No. JCYJ20210324120 \\ 012033 and JSGG20210802154402007), Shenzhen Science and Technology Program (Grant No. WDZC20231128091437002), the Major Key Project of PCL for Experiments and Applications (PCL2021A06), and Overseas Cooperation Research Fund of Tsinghua Shenzhen International Graduate School (HW2021008).

\section*{Declaration of competing interest}
The authors declare that they have no known competing financial interests or personal relationships that could have appeared to influence the work reported in this paper.

\section*{CRediT authorship contribution statement}
Yinghui Li: Conceptualization,  Formal analysis, Investigation, Methodology, Writing - original draft.

Shirong Ma: Conceptualization,  Formal analysis, Software, Methodology, Writing - original draft.

Shaoshen Chen: Formal analysis, Validation, Visualization, Writing - original draft.

Haojing Huang: Investigation, Data curation, Writing - review \& editing

Shulin Huang: Resources, Writing - review \& editing

Yangning Li: Writing - review \& editing

Hai-Tao Zheng: Funding acquisition, Supervision, Writing - review \& editing

Ying Shen: Project administration, Resources, Writing - review \& editing

\section*{Data availability}
The data used in our work will be made available on request.




\bibliographystyle{elsarticle-num-names} 

\bibliography{IEEEabrv}






\end{document}